# Completing point cloud from few points by Wasserstein GAN and Transformers

Xianfeng Wu[1, 2, 4, 5, 6#], Jinhui Qian[1, 4, 6#], Qing Wei[1, 4, 6#], Xianzu Wu[2, 3#], Xinyi Liu[4], Luxin Hu[1, 4, 6], Yanli Gong[1, 4, 6], Zhongyuan Lai[1, 4, 5, 6*], Libing Wu[2*]

[1]State Key Laboratory of Precision Blasting, Jianghan University, Wuhan 430056, P.R. China
[2]School of Cyber Science and Engineering, Wuhan University, Wuhan, 430072, P.R. China
[3]College of Geophysics and Petroleum Resources, Yangtze University, Wuhan 430100, P.R. China.
[4]School of Artificial Intelligence, Jianghan University, Wuhan 430056, P.R. China
[5]Bingling Honorary School, Jianghan University, Wuhan 430056, P.R. China
[6]Institute for Interdisciplinary Research, Jianghan University, Wuhan 430056, P.R. China
laizhy@jhun.edu.cn*
wu@whu.edu.cn*

## Abstract

*In many vision and robotics applications, it is common that the captured objects are represented by very few points. Most of the existing completion methods are designed for partial point clouds with many points, and they perform poorly or even fail completely in the case of few points. However, due to the lack of detail information, completing objects from few points faces a huge challenge. Inspired by the successful applications of GAN and Transformers in the image-based vision task, we introduce GAN and Transformer techniques to address the above problem. Firstly, the end-to-end encoder-decoder network with Transformers and the Wasserstein GAN with Transformer are pre-trained, and then the overall network is fine-tuned. Experimental results on the ShapeNet dataset show that our method can not only improve the completion performance for many input points, but also keep stable for few input points. Our source code is available at https://github.com/WxfQjh/Stability-point-recovery.git.*

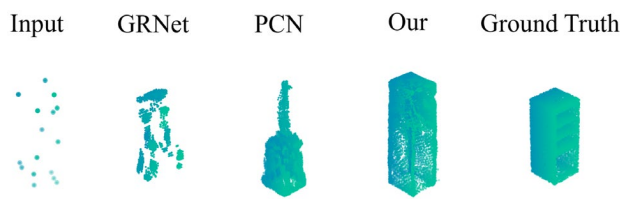

Figure 1. The extreme case i.e., 16 points of partial point cloud input for dense point cloud complementation, PCN and GRNet are identified wrongly as Lamp and Chair respectively.

## 1. Introduction

Data completion is a core technology for many vision and robotics applications. With the development of deep neural networks, the performance of data completion is steadily improving and showing increasingly promising applications in autonomous driving, robotics, and augmented reality. Deep neural networks were first successful in image restoration due to the natural orderliness, uniformity, and regularity of image structures. Compared with image inputs, although 3D point clouds have the advantage of higher dimensional spatial information and are less susceptible to light variations, their disorder, non-uniformity, and irregularity make it not easy to design a completion network that directly uses 3D point clouds as input. PCN [1] is one of the pioneers of three-dimensional point cloud as input. The network considers displacement invariance and introduces two stacked PointNet [2] layers and multi-stage point cloud-generated network in the design. It not only achieves initial success in point cloud completion, but also has strong stability against the number changes of input points. The subsequent improvement methods [3-8] are mostly aimed at the point cloud completion in the case of many input points to improve the completion effect. However, when the number of input points changes, especially when it is very small, it will often cause drastic changes in the neighborhood or even disappear, thus reducing the effect of point cloud completion. So, these improvements are mostly at the expense of stability.

In the practical application of point cloud completion, the stability is often more important than the improvement of the effect. Here are two typical application scenarios: First, in autonomous driving applications, the collected point clouds are often very sparse due to the influence of radar sensor resolution and occlusion, in which case the effect of completion will directly affect the recognition of targets. Second, in robot vision, the points used for small targets are often very few, and in this case, the stability of the method is crucial for the detection of subsequent targets.

For this reason, we expect to propose a point cloud

# Equal contribution

completion *method* that not only performs comparably to the best existing methods in the case of a relatively large number of points, but also maintains the completion effect well in the case of very few points. GAN [9] and Transformer [10] techniques have achieved very good results in the recovery of images. It is shown that the weights of the feature extraction network are biased due to data residuals, and the GAN network can correct the weights well by introducing discriminators to get features closer to the true value of the object. And Transformer has a unique advantage in maintaining stable performance by introducing a global self-attentive mechanism. Inspired by the above work, we also try to introduce GAN and Transformer into PCN, firstly, we introduce Transformer [11] module in the encoder network of PCN to improve the stability of feature extraction; Then, PointNet++ [12] and Transformer [13] modules are introduced into PCN decoder network to improve the generation effect of dense point cloud by making full use of multi-level neighborhood features. Finally, WGAN [14] network is inserted between encoder and decoder to further optimize the network weights.

Since we use WGAN and Transformer in the feature extraction process to improve the completion accuracy and stability of the network, the proposed method is not only better than PCN in the case of many points, but also more stable in the case of few points. This is partially confirmed by the demonstration experiments shown in Fig. 1. Further, the completion results on the ShapeNet dataset [15] show that the proposed method outperforms PCN for a typical input of 2048 points, while for 16 points, the proposed method still maintains better completion for most objects, while the PCN method either generates The PCN method either generates the wrong class of objects or fails completely, thus fully verifying the effectiveness of our method.

The main contributions of this paper are as follows:

1. We introduce Transformer technology based on PCN, which does not rely on neighborhood information and therefore has good stability for the case of domain disappearance caused by too sparse input points

2. We also introduce the WGAN technique, which can further enhance the stability of the extracted features by introducing discriminators that can well bridge the difference in distribution between the input features and the true-value features caused by too sparse input points.

3. The experimental results on the ShapeNet dataset show that it is not only effective in improving the PCN completion effect in the case of many points, but more importantly, the completion effect remains stable in the case of a small number of points.

The remainder of this paper is organized as follows: Section 2 presents the related work, Section 3 gives the network approach based on WGAN and Transformer, Section 4 gives the experimental results, and Section 5 gives the summary and outlook of this paper.

## 2. Related work

Compared with images, 3D points clouds have richer spatial information and are less susceptible to illumination changes, which have unique advantages in pattern recognition. However, the natural disorder, nonuniformity and irregularity of 3D point clouds increase the difficulty of designing 3D point completion methods. In the following, two types of typical 3D point completion methods are introduced: voxel-based methods and point-based methods, followed by the application of GAN and Transformer techniques on images.

### 2.1. Voxel-based methods

The voxel-based method first voxelates the input residual point cloud and then completes it. GRNet [16] is one of the representative methods. This method inputs the voxels of the mutilated objects into the 3D CNN to obtain the complementary results of the voxel representation, and then obtains the complementary results of the point cloud representation by MLPs. Since the 3D CNN computation presupposes the existence of voxel neighborhoods, the existence of voxel neighborhoods can be guaranteed when the number of input points is large, but when the number of input points is small, the voxel neighborhoods will disappear, which in turn causes the method to fail.

### 2.2. Point-based method

Unlike voxel-based methods, point-based methods perform point cloud completion directly on the vestigial points, and PCN [1] is one of the pioneers. The method is divided into two parts, encoder, and decoder. The encoder extracts global features after PointNet [2], while the decoder is divided into two stages, the first stage generates a coarse point cloud through a fully connected network, and the second stage generates a fine point cloud through a FoldingNet [3] network. Since no neighborhood calculation is involved in the whole completion process, it has a certain stability for the change of the number of points in the point cloud. The disadvantage is that the completion effect is average.

For this reason, many improvement methods have been proposed, such as PF-Net [4], which obtains point clouds with more prominent geometric features by layer-by-layer downsampling, so that the coding end is more focused on these more representative point clouds, and TopNet [5] decoder, which progressively increases the resolution of the generated point clouds through a tree-like structure to obtain more accurate completion results, and SA-Net [6], which follows the example of PointNet++, which extracts point clouds according to SA-Net extracts features from high to low resolution layer by layer, and then generates features from low to high resolution layer by layer at the decoding end, and finally outputs point cloud coordinates. Recently, PMP-Net [7] and its enhanced version, PMP-Net++ [8] method, have been proposed. Unlike the generative methods mentioned above, both methods obtain the completion results by gradually moving each point in the input stubby point cloud. Among them, the coding side follows the design idea of PointNet++ [12] to obtain hierarchical features involved in the calculation of the move step according to the resolution from high to low. Since the above-mentioned improved methods either involve downsampling or neighborhood calculation, they are all difficult to apply to the case where the number of input residual points is small.

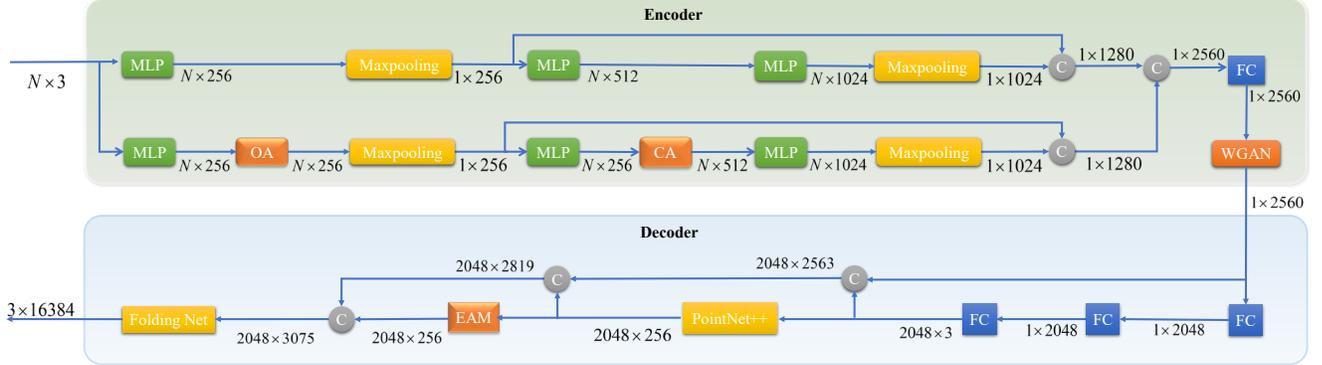

Figure 2. Overall architecture. The partial point cloud is first input to the encoder to extract the multi-level global feature information of the point cloud, and then input to the WGAN network which processes the multi-dimensional global features, and regenerate the new multi-dimensional global features by the Generator in WGAN, and input the ground truth point cloud to the encoder to obtain the multi-dimensional global features of the real point cloud, and then generate the new multi-dimensional global features by the same process. Discriminator determines whether it is the generated multidimensional global feature, and then generates the complemented point cloud by the decoder.

### 2.3. GAN

Generative Adversarial Networks (GAN) [9] first achieved great success in image generation. A typical GAN network consists of a generator and a discriminator, which play against each other to make the distribution of the generated image samples as close as possible to the distribution of real image samples, and finally achieve a false image generation. Among many GAN networks, WGAN [14] achieves the overlap of the two distributions by introducing noise into the generated image samples and the real image samples, and thus has strong stability in the training process, and has been widely used in vision tasks with image as input.

### 2.4. Transformer Technique

Transformer techniques first made a splash in the fields of machine translation and natural language processing [17]. Inspired by this, a large amount of work [9, 18] has been done to introduce the Transformer technique to vision tasks with image as input, which has also been successful.

The self-attentive mechanism in Transformer is uniquely advantageous in keeping the performance of vision tasks stable due to its global relevance [19-20]. At the same time, the self-attentive mechanism is essentially an ensemble operation, and its results are not affected by changes in the size of the alignment and input ensembles, making it ideal for processing point cloud data that are essentially the same ensemble [11, 21].

Inspired by the above work, our main idea is to combine PCN with WGAN [14] and Transformer [11, 13] to improve the stability of PCN in the case of very small number of points. Our approach is described in detail below.

## 3. Our method

### 3.1. Overall Architecture

In this paper, we design a codec network based on Transformer with WGAN for point completion, as shown in Fig. 2. First, the encoder is pre-trained with the real point cloud. Then, the partial point cloud and the completion point cloud are put into the pre-trained encoder-decoder network to obtain the noisy multi-dimensional global features and the perfect multi-dimensional global features respectively, and the perfect multi-dimensional global features are used as real values and the noisy multidimensional global features are used as fake for training, and finally, the pre-trained codec and the Generator in GAN network synthesizes the overall network framework fine-tune hyperparameters. When the residual and real point clouds are fed into the twin encoder (Encoder) to obtain the multidimensional global feature vector $MGFV_p$, the feature vector is relatively noisy. Then, this paper proposes a WGAN-GP-based method, including Generator and Discriminator modules, which aims to establish the connection between the multidimensional global feature vectors of the residual and real point clouds. The regenerated vector $G(MGFV_p)$ obtained by Generator is provided to Decoder for completing the sparse point clouds to generate dense point clouds.

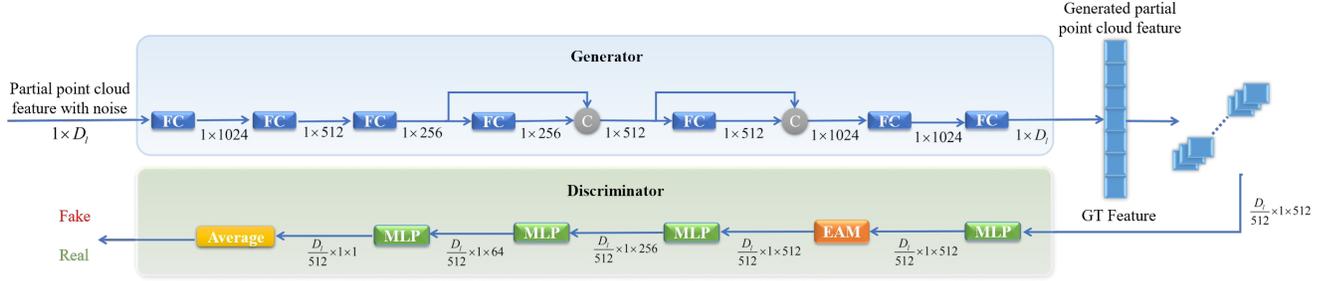

Figure 3: GAN network framework for processing multidimensional global features. partial point cloud generates noisy multidimensional global features by pre-trained encoders, generates processed multidimensional global features by pyramid-like Generator network, and the perfect multidimensional global features generated by Completion point cloud and the above processed multidimensional global features are fed into the fully connected network with external-attention to identify whether they are generated multidimensional global features. The perfect multidimensional global features generated by the Completion point cloud and the above processed multidimensional global features are input to the fully connected network with external-attention to identify whether they are the generated multidimensional global features. The dotted line is only for data transfer or reshape.

### 3.2. Ensemble Encoder with Transformers

In this paper, we design an integrated encoder, the encoder $Encoder_P$ is based on PointNet multi-level global feature extraction network, and the encoder $Encoder_T$ is based on PointNet multi-dimensional feature extraction network with Transformer, synthesized into an integrated encoder to extract multi-level global feature information of point clouds. The specific design is shown in Figure 2. The multi-level global feature information obtained from each encoder section is as follows.

$$MGFV_{P-PN} = PN-CMLP(P)$$
$$MGFV_{P-T} = T-CMLP(P)$$

Among them, PN-CMLP is a PointNet-based combination MLP, T-CMLP is a PointNet-based combination MLP with Transformer, and P is the input point cloud.

The final output multi-level global feature information $MGFV_P$ is as follows.

$$MGFV_P = LeakyReLU\left(FC\left(concat\left(MGFV_{P-PN}, MGFV_{P-T}\right)\right)\right)$$

where FC is Fully connected Layer.

### 3.3. Transformer-based GAN module

Inspired by the great success of GAN in image editing work, we migrate the learning to the point completion task in this paper, we design an external-attention based WGAN module to correct noisy multidimensional global features to perfect multidimensional global features. The Generator part takes a simple fully-connected layer, and the Discriminator takes a fully-connected network with external-attention to identify whether the multidimensional global features are noisy or not. The specific design is shown in Fig. 3.

### 3.4. PointNet++ based decoder module

The decoder of PCN only generates the coarse point cloud from the global features through the fully connected network and then generates the dense point cloud through the folding net layer, so the randomness is too large, and the recognition effect is not optimal. So, we extract the local features of the generated coarse point cloud through the PointNet++ feature extraction layer after generating the coarse point cloud through the fully connected network and concatenate the global features and the generated coarse point cloud in the input decoder and then generate the final dense point cloud through the folding net layer. The decoder framework is shown in Fig. 2.

### 3.5. Loss function

**The Encoder-Decoder loss function**. For both the pre-trained network and the overall network consisting of Ensemble Encoder and PointNet++-based Decoder, i.e., the same loss function as the current point completion task, Chamfer Distance (CD) and Earth Mover's Distance (EMD). In this paper, we briefly review CD and EMD.

Among them, CD and EMD are the most used forms of loss function calculation in 3D point completion techniques.

Since both point clouds are disordered, the loss is invariant to the alignment of the point cloud. Two candidate alignment invariant functions are introduced: Chamfer distance (CD) and Earth Mover distance (EMD) [22]

$$CD(S_1, S_2) = \frac{1}{|S_1|}\sum_{x \in S_1} \min_{y \in S_2} \|x-y\|_2 + \frac{1}{|S_2|}\sum_{y \in S_2} \min_{x \in S_1} \|y-x\|_2$$

where CD calculates the average nearest point distance between the output point cloud S1 and the ground truth point cloud S2. A symmetric form of CD is used, where the first term forces the output points to be close to the ground truth points and the second term ensures that the ground truth point cloud is covered by the output point cloud. It is not necessary to calculate CD so that S1 and S2 are of the where EMD finds a bijective function φ : S1 → S2 that minimizes the average distance between the corresponding points. In practice, finding the best φ is very expensive, so an iterative $(1+\varepsilon)$ approximation scheme is used [23]. Unlike the calculation of CD, EMD requires that S1 and S2 have the same magnitude.

The loss function consists of two $d_1$ and $d_2$, weighted by the hyperparameter $\alpha$. The first term is a coarse output of the distance between $Y_{coarse}$ and the subsampled ground truth $Y_{gt}$ of the same size as $Y_{coarse}$. The second term is a detailed output of the distance between $Y_{detail}$ and the full ground truth $Y_{gt}$.

$$L\left(Y_{coarse}, Y_{detail}, Y_{gt}\right) = d_1\left(Y_{coarse}, \tilde{Y}_{gt}\right) + \alpha d_2\left(Y_{detail}, Y_{gt}\right)$$

**WGAN loss function**. The training of Generator G and Discriminator D is based on the multidimensional global feature vector space of the point cloud. Let P denote the Partial point cloud, $x = Encoder(P)$ denote the multidimensional global feature vector of the Partial point cloud, Y denote the ground truth, and $y = Encoder(Y)$ denote the multidimensional global feature vector of the ground truth. First, we introduce the training of Discriminator D. Here we adopt the loss of Discriminator from the literature [WGAN-GP] as the loss of our network. the loss function of Discriminator Loss is as follows:

$$L = \underbrace{\mathbb{E}_{\tilde{x} \sim \mathbb{P}_g}[D(\tilde{x})] - \mathbb{E}_{x \sim \mathbb{P}_r}[D(x)]}_{\text{critic loss}} + \underbrace{\lambda \mathbb{E}_{\hat{x} \sim \mathbb{P}_{\hat{x}}}\left[\left(\nabla_{\hat{x}} D(\hat{x})_2 - 1\right)^2\right]}_{\text{gradient penalty}}$$

where $\mathbb{P}_g$ is the multidimensional global feature vector space generated by Generator G for the multidimensional global features of the partial point cloud, $\mathbb{P}_r$ is the multidimensional global feature vector space of the completion point cloud, $\mathbb{P}_{\hat{x}}$ is the noisy multidimensional global feature vector space formed by the multidimensional global features of the completion point cloud and the partial point cloud, and is the noisy multidimensional global feature vector space formed by the multidimensional global features of completion point cloud and the multidimensional global features of partial point cloud as penalty terms.

The training of Generator G also requires fixing the parameters of Discriminator D. The loss function Loss $G_{adv}$ is as follows:

$$LossG_{adv} = L_{bce}(D(x), 1) + L_{bce}(D(G(x)), 1)$$

where bce is the Binary Cross Entropy with the following equation.

same size.

$$EMD(S_1, S_2) = \min_{\phi: S_1 \to S_2} \frac{1}{|S_1|} \sum_{x \in S_1} \| x - \phi(x) \|_2$$

$$L_{bce}(z, t) = (t \log(z) + (1 - t) \log(1 - z))$$

Minimizing this loss means that Generator G will make Discriminator D as "confusing" as possible, i.e., D will not be able to correctly distinguish between sample sources. However, minimizing this loss alone may lead to instability in the actual training process. We apply the L1 distance between the y generated by G and the Generator-corrected x to the loss function. the L1 loss function LossL1 and the final loss function LossG are given below:

$$Loss_{L_1} = \| y - G(x) \|_1$$

$$LossG = \alpha \times LossG_{adv} + \beta \times Loss_{L_1}$$

## 4. Experiments

To verify that our method can not only improve the completion performance in the case of many points, but also maintain the stability of the completion results in the case of few points, we design a series of experiments. In Section 4.1, we give the common datasets and training models. In Section 4.2, we give the completion results for the case of many points, in Section 4.3, we give the completion results for the case of few points, and in Section 4.4, we give the results of the ablation experiments.

### 4.1. Data and Model Training

We use ShapeNet as our experimental dataset, which contains 8 classes, namely airplane, cabinet, car, chair, lamp, sofa, table, and vessels, with a total of 30,974 samples. The dataset is divided into two parts, where 28,974 samples are used for training, 800 samples for validation, and 1,200 samples for testing. For each training sample, a partial point cloud of the eight different views to which it corresponds is included. For each validation sample and test sample, only the corresponding point cloud of its corresponding single viewpoint is included. The pair contains one sample point cloud and eight corresponding residual point clouds.

For model training, we first pre-train the overall network consisting of encoders and decoders end-to-end, then pre-train the encoders in combination with the WGAN network, and finally fine-tune the encoder WGAN network and decoder network end-to-end. All our models are trained with Adam optimizer both for pre-training and fine-tuning, with an initial learning rate of 0.0001, 250 epochs, and a batch size of 32. Every 20 epochs, the learning rate decays by 0.7.

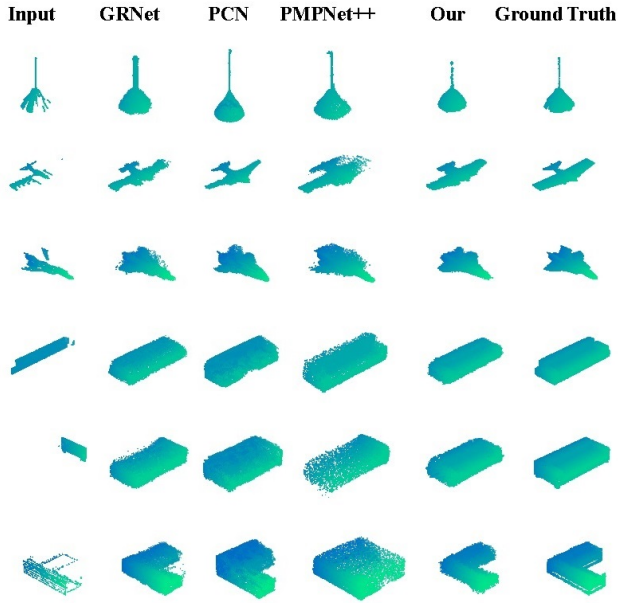 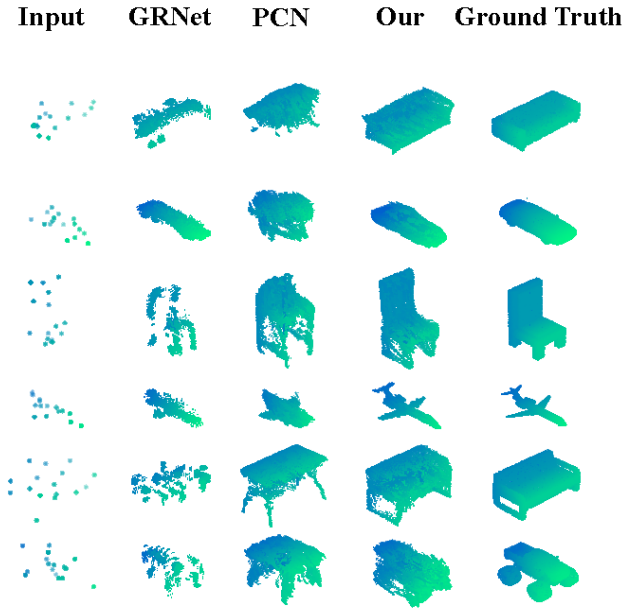

Figure 4. Qualitative results completed from 2048 points.

Figure 5. Qualitative results completed from 16 points.

## 4.2. Fine point completion

To verify that our method can effectively improve the point cloud completion with a larger number of points, we use a partial point cloud of 2048 points as input, as in the literature PCN. Also, we selected PCN and GRNet as our comparison objects.

The following figure shows the completion results of PCN, GRNet and our method when the input number of points are 2048. Compared with PCN, our method is closer to the true value in appearance, and compared with GRNet, our method produces fewer noise points, which indicates that our method is indeed effective in improving the point cloud completion.

Table I shows the average reduction rates of CD and EMD between the completion results and the true values for PCN, GRNet and our method on all eight classes of objects. Using our method, the average reduction rate of method i is defined as

$$CD \downarrow = (CD_i - CD_{our}) / CD_i,$$
$$EMD \downarrow = (EMD_i - EMD_{our}) / EMD_i,$$

Our method reduces the distance between the completion result and the true value on most classes. Compared to PCN, at most 17.05% CD and 9.76 % EMD are reduced. Compared to GRNet and PMPNet++, it reduces the CD by at most 22.75% and 34.17% respectively. On average, the 7.95%CD and 1.69%EMD are reduced compared to PCN and 4.33%CD and 4.05%CD compared to GRNet and PMPNet++ respectively. This fully demonstrates that our method can effectively improve the performance of completion in the case of many points, which in turn facilitates the improvement of the performance of subsequent applications such as recognition and detection.

## 4.3. Sparse point cloud completion

To verify that our method can maintain the stability of the completion effect even with a small number of points, we choose two representative methods PCN and GRNet as the comparison objects. We randomly sampled part of the point cloud used for testing to 16 points and then input them into the above trained model. The completion results are shown as follows: the PCN method either generates unknown objects, as shown in the first two rows of Fig. 5., or the completion results are very different from the true values, as shown in the middle two rows, or completes to other classes of objects, as shown in the last two rows; the completion results of the GRNet method contain a large number of fragments, which makes it difficult to form a connected object; while except for the last row, our method can generate objects with a small number of points Except for the last row, our method is able to generate visually close and recognizable results with the true value even with a small number of points, which fully demonstrates that our method can maintain the stability of the completion effect even with a small number of points.

As shown in Table II, our method reduces the distance between the complement result and the true value on most classes. Compared to PCN, at most 35.71% CD distance and 27.59% EMD distance are reduced. Compared to GRNet, it reduces the CD distance by at most 57.73%. On average, the 24.71%CD distance and 18.56%EMD distance are reduced compared to PCN and 46.26%CD distance compared to GRNet. This fully demonstrates that our method can effectively improve the performance of complementation in the case of many points, which in turn facilitates the improvement of the performance of subsequent applications such as recognition and detection.

Table I. The average reduction rate of our method comparing with PCN, GRNet and PMPNet++ when the number of input points is 2048.

| ShapeNet | | Airplane | Cabinet | Car | Chair | Lamp | Sofa | Table | vessel | Average |
|---|---|---|---|---|---|---|---|---|---|---|
| PCN | CD↓ | 5.11% | 0.72% | -6.68% | 12.00% | 8.45% | 21.55% | -2.64% | 17.05% | 7.95% |
| | EMD↓ | 5.26% | 3.20% | -9.62% | -1.64% | 9.76% | 5.08% | -12.73% | 5.00% | 1.69% |
| GRNet | CD↓ | 22.75% | 9.73% | 1.10% | 6.47% | -27.45% | 19.37% | -0.10% | -1.59% | 4.33% |
| PMPNet++ | CD↓ | 16.49% | 13.14% | 7.39% | 7.47% | 34.17% | 17.91% | -0.62% | -2.29% | 4.05% |

Table II. The average reduction rate of our method comparing with PCN and GRNet when the number of input points is 16.

| ShapeNet | | Airplane | Cabinet | Car | Chair | Lamp | Sofa | Table | vessel | Average |
|---|---|---|---|---|---|---|---|---|---|---|
| PCN | CD↓ | 35.37% | 24.36% | 35.71% | 20.54% | 21.93% | 12.85% | 30.47% | 23.88% | 24.71% |
| | EMD↓ | 23.73% | 20.16% | 27.59% | 8.04% | 11.02% | 9.47% | 11.00% | 17.65% | 18.56% |
| GRNet | CD↓ | 54.73% | 52.00% | 54.99% | 38.51% | 30.69% | 41.04% | 57.73% | 39.83% | 46.26% |

Table III. The average reduction rate of the combinations of PCN with various models comparing with PCN when the number of input points is 2048

| ShapeNet | | Airplane | Cabinet | Car | Chair | Lamp | Sofa | Table | vessel | Average |
|---|---|---|---|---|---|---|---|---|---|---|
| PCN + PointNet++ module | CD↓ | 1.19% | 1.35% | -7.44% | 1.52% | 2.90% | 9.61% | -2.22% | 6.49% | 2.19% |
| | EMD↓ | 0.00% | 1.56% | -3.85% | 1.64% | 3.66% | 0.00% | -7.27% | 0.00% | 0.00% |
| PCN + Transformer | CD↓ | 1.70% | -0.90% | -9.16% | 6.17% | 5.87% | 17.27% | -1.27% | 7.97% | 4.36% |
| | EMD↓ | 0.00% | -1.56% | -13.46% | -3.28% | 7.32% | 1.69% | -7.27% | 1.67% | 0.00% |
| PCN + WGAN | CD↓ | 3.75% | 0.18% | 1.19% | 7.86% | 6.10% | 17.79% | 0.42% | 7.23% | 6.25% |
| | EMD↓ | 0.00% | 3.13% | 0.00% | 0.00% | 7.32% | 0.00% | 0.00% | 0.00% | 1.69% |
| Our | CD↓ | 5.11% | 0.72% | -6.68% | 12.00% | 8.45% | 21.55% | -2.64% | 17.05% | 7.95% |
| | EMD↓ | 5.26% | 7.81% | -9.62% | -1.64% | 9.76% | 5.08% | -12.73% | 5.00% | 1.69% |

Table IV. The average reduction rate of the combinations of PCN with various models comparing with PCN when the number of input points is 16

| ShapeNet | | Airplane | Cabinet | Car | Chair | Lamp | Sofa | Table | vessel | Average |
|---|---|---|---|---|---|---|---|---|---|---|
| PCN + PointNet++ module | CD↓ | 4.53% | 1.67% | 4.00% | 3.18% | 0.86% | 4.33% | 4.08% | 5.04% | 3.33% |
| | EMD↓ | 3.39% | 0.81% | 1.15% | 1.79% | 0.00% | 1.05% | 1.00% | 1.18% | 1.03% |
| PCN + Transformer | CD↓ | 13.41% | 5.78% | 10.32% | 5.63% | 5.87% | 9.43% | 8.17% | 8.17% | 7.94% |
| | EMD↓ | 5.08% | 1.61% | 1.15% | 2.68% | 1.69% | 2.11% | 2.00% | 2.35% | 2.06% |
| PCN + WGAN | CD↓ | 27.87% | 14.68% | 25.49% | 14.52% | 11.96% | 3.97% | 19.01% | 13.39% | 15.48% |
| | EMD↓ | 16.95% | -21.77% | 25.29% | 7.14% | 8.47% | 6.32% | 6.00% | 7.06% | 5.15% |
| Our | CD↓ | 35.37% | 24.36% | 35.71% | 20.54% | 21.93% | 12.85% | 30.47% | 23.88% | 24.71% |
| | EMD↓ | 23.73% | -20.16% | 27.58% | 8.04% | 11.02% | 9.47% | 11.00% | 17.65% | 8.25% |

### 4.4. Ablation Study

Tables III and IV give the average reduction rate of the CD and EMD between the PCN with the help of each module for all eight categories of completion results and the ground truth when the numbers of input points are 2048 and 16, respectiely. Here, the average reduction rate for module i (i = transformers, PointNet++ or WGAN) is defined as

$$CD\downarrow = (CD_{PCN} - CD_i)/CD_{PCN},$$
$$EMD\downarrow = (EMD_{PCN} - EMD_i)/EMD_{PCN},$$

It can be seen that when the number of input points is 2048, with the help of PointNet++, transformers or WGAN, CD drops at most 9.61%, 17.27%, 17.79%, average drops 2.19%, 4.36%, 6.25%, EMD drops at most 3.66%, 7.32%, on average drop 0.00%, 0.00%, 1.69%, thus making the overall CD and EMD of our method drop at most 21.55% and 9.76%, on average 7.95%, 1.69%. And when the number of input points is 16, with the help of PointNet++, transformers or WGAN, CD drops at most 5.04%, 13.41%, 27.87%, average drops 3.33%, 7.94%, 15.48%, EMD drops at most 3.39%, 5.08%, 25.29%, drop on average 1.03%, 2.06%, 5.15% and thus make the overall CD and EMD of our method drop at most 35.71% and 27.58%, and drop 24.71% and 8.52% on average. This not only shows that each module can effectively reduce the distance between the completion result and the true value, and improve the completion effect, but also in the extreme case where the number of points is small, its contribution to the completion effect is greatly improved. It further reflects the contribution of each module in maintaining the stability of the model.

### 5. Conclusion

In practical computer vision and robotics applications, small and weak objects often need to be detected and recognized. These objects are often represented by very few points, and the effectiveness of their completion directly affects the performance of subsequent detection and recognition. However, most of the existing methods perform completion for objects with many point representations, and they often have poor completion effects with few points, which poses great difficulties for subsequent detection and recognition. For this reason, we propose a point cloud completion method for the case of few points. Based on PCN, we introduce WGAN to modify the weights of the global feature extraction network to reduce the impact on global feature extraction due to the bias of feature distribution caused by sparse point clouds. Meanwhile, we introduce the Transformer module to further improve the stability of feature extraction. The experimental results on ShapeNet show that our method not only improves the effect of point cloud completion in the case of many points, but also maintains the stability of the completion results in the case of few points, and thus has stronger adaptability in the complex and changing practical application environment.